# Precision Improvement of MEMS Gyros for Indoor Mobile Robots with Horizontal Motion Inspired by Methods of TRIZ

Dongmyoung Shin[*], Sung Gil Park[*], Byung Soo Song[#], Eung Su Kim[#], Oleg Kupervasser[#1], Denis Pivovartchuk[#], Ilya Gartseev[#], Oleg Antipov[#], Evgeniy Kruchenkov[#], Alexey Milovanov[#], Andrey Kochetov[#], Igor Sazonov[#], Igor Nogtev[#], and Sun Woo Hyun[#]

[*]LG Electronics, Korea
[#]LG Technology Center of Moscow, Russia
[1]olegkup@yahoo.com

*Abstract* − **In the paper, the problem of precision improvement for the MEMS gyrosensors on indoor robots with horizontal motion is solved by methods of TRIZ ("the theory of inventive problem solving").**

## I. Introduction

The last years, indoor robots appear widely our life. There is big necessity for mass production of cheap and qualitative sensors being "sense organs" of the robots. To the first applicants for this role are MEMS sensors - cheap, simple for production, having the small size [1]. Unfortunately, they have also serious weakness. Namely, it is low precision. It is possible to compensate this weakness by computer processing of output signal of sensors. However, such processing is a complex algorithmic problem. We take a special case of this problem - precision improvement for the MEMS gyrosensors on indoor robots with horizontal motion. We found the problem's solution with application of TRIZ ("the theory of inventive problem solving") tools [2].

## II. State of the Art of the Problem

The modern MEMS (Micro-electro-mechanical-system) gyroscopes represent one of achievements of the modern technique and are widely used for indoor robots with horizontal motion. The gyroscope is motionlessly attached on a robot with vertical direction of the gyroscope's axis Z (Fig. 1).

Small, flat, and cheap sensors allow to measure angular velocities. And on the basis of the measure, it is possible to get angles of rotation by integration. Unfortunately, they have also serious weaknesses. "Zero drift" is the main one. It means that even at zero angular velocity the sensor gives non-zero indications. This constant deviation ("zero shift") can be excluded by deduction of the correspondent constant (which is measured by the sensor before the beginning of a motion). However, the problem exists to do it because of "drifts" (i.e. changes) of the "zero shift". This "drift of zero" influences not strongly measuring of angular velocity. However, "drift of zero" strongly affects integration of angular velocities, because of cumulative property of integration. It is possible, complicating a construction of the sensor, considerably to reduce the "drift of zero". Nevertheless, it leads to the considerable rise in price of the sensor.

Suppose that the interior structure of a gyroscope is not known and is "black box". It is possible to a minicomputer, allowing to transform output signal of the gyroscope. How to improve the precision of the sensor without changing the interior structure of a gyroscope (and, consequently, without complicating the sensor) by using the minicomputer only?

## III. The solution by the TRIZ method (ARIZ-85 ("the algorithm of inventive problem solving"-85) steps)

### A. The analysis of an initial situation

Definition of the final goal for the problem solution

*1)* The final goal is measuring an angle of rotation with the precision, which much exceeds much more effect, obtained by direct integration of the measured angular velocity.

*2)* For the solution of the problem, it is impossible to change the interior structure of a gyroscope. The structure is not known and it is "black box". It is impossible fusion with some other sensors. The gyroscope motionlessly attached on a robot with vertical direction of the gyroscope's axis Z. Thus the sensor is used for indoor robots with horizontal motion

*3)* Quantity indicator of effect is the measurement

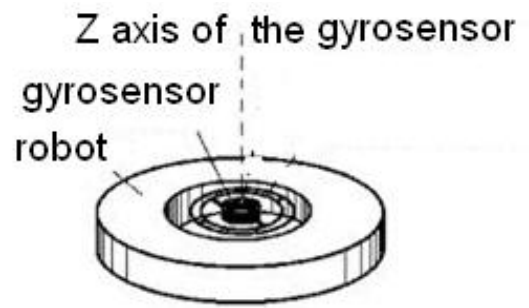

Fig. 1.  Gyroscope setting on the robot.

accuracy of a rotation angle after 10 minutes of motion for the set of test motions.

*B. The analysis of the problem*

*1)* We write be low a requirement of a mini-problem.

The technical system for measuring a rotation angle includes
a. The system is a gyroscope with the power supply system, capable to measure three angular velocities on three axes (axis Z is guided vertically), a temperature sensing device, the built - in minicomputer for processing a output signal
b. The interior structure of a gyroscope is a subsystem. The subsystem is forbidden for changing. There is also no model of a subsystem – It is "black box".
c. Supersystem of the sensor is the indoor robot with horizontal motion. The indoor robot consists of carcass, engine, chassis, fan, system of dust taking, the sensors of walls
d. Global supersystem is room with floor, walls, ceiling, furniture, air

Our purpose is to receive a precise rotation angle without any interior changes to the sensor. This purpose is easily achievable by introduction of a minicomputer: the minicomputer processes an output signal to clean "noise" without any interior changes to the sensor. Nevertheless, there is a following problem:

At an output of a gyroscope is a <u>sum</u> of the exact signal (the angular velocity) and the noise (which perturbates the exact signal). I.e. from the single source (the output) we need to receive the information about two quantities, i.e. it is necessary to find two unknown variables from single equation. We want to find and to clean the noise from the output signal. However, it is not clear, where the noise begins and where the exact signal comes. Only one method exists to do it: to find some additional information about the motion. It would give us the additional second equation and would let us to find the problem's solution.

Additional "source" of the information about motion can be some other sensor as it is recommended in [4-6].

As additional "source" of the information about motion, we can use also fast rotation at $\pm 180^0$ of the sensor as it is recommended in [7]. Indeed, thus the direction of the measured angular velocity with respect to axis Z of gyrosensor varies on opposite. However, the noise of the gyrosensor remains unchanged and directed along axis Z. Consequently, by this way we sum two same useful signals and subtract two unknown signals (noises) with opposite sign and the same absolute values.

However, both these methods lead to complication and the price increase of the sensor.

Thus, we can get:

<u>Technical contradiction</u> **(TC)**: we want to take out the noise from the output signal by the minicomputer. We have only one source of the information for two variables –it is the output signal of a gyroscope. It is not enough information for getting both the useful signal and the noise. By introducing some additional "source" of the information, we complicate system and raise the price of the system.

We can reformulate this technical contradiction in two forms:

**TC1:** we insert additional "source" of the information, but we complicate system.

**TC2:** we do not insert additional "source" of the information and we do not complicate thus system but then we have a big error of the useful signal because we do not take out the noise.

*2)* It is necessary to take out the noise; otherwise, we have no solution. Therefore, we select **TC1** as more appropriate contradiction.

*3)* In this case, "product" is angular velocity, found by the minicomputer from the output. "Tool" is the minicomputer, separating the noise and the useful signal from the output signal. "Treated object", which is treated by the "tool" is the output signal. This signal is a source of the measured information about motion.

*C. The analysis of the problem model*

*1)* The graphic representation of the Technical Contradiction is given on Fig. 2.

*2)* Transition to **PC1** (to the physical contradiction).

Let us formulate the Physical Contradiction:

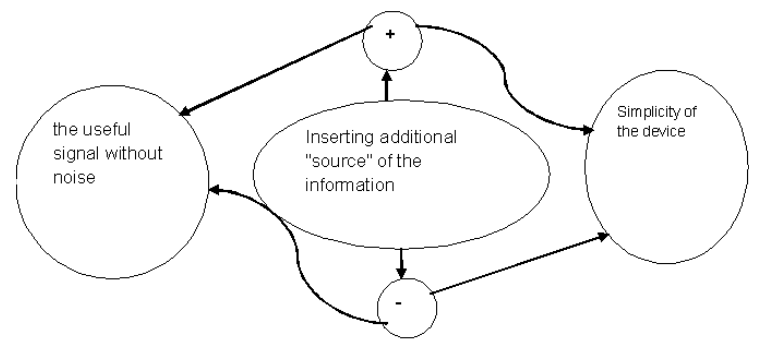

a)

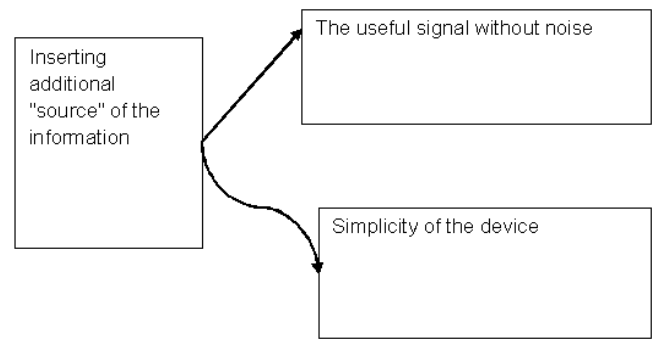

b)

Fig. 2. Gyroscope setting on the robot.

**PC1:** the measuring system is complex, because additional "source" of the information is necessary for extraction of noise for us. At the same time, the measuring system is simple, because we do not use additional "source" of the information.

*3)* The ideal final result

We get the additional information, which is necessary to find the noise. However, we do not insert the additional source of the information, which can complicate the system.

*D. The resolution of the Physical Contradiction*

*1)* Separation the inconsistent properties over time.

We can once rotate a gyroscope with some known angular velocity and then to find precise deviation (from the angular velocity) the by subtraction of the known angular velocity from the output signal. In future, if we measure the same output signal of the gyroscope we can assume that deviation is the same and find the useful signal. This method is named "calibration". During the calibration we can set a wide set of known angular velocities, then find the correspondent deviations for these velocities, and finally to draw up the table with correspondence between known angular velocities and measured output signals [8-11].

However, are these deviations always identical for the same angular velocities? The answer - "No, it is not". Actually, these deviations depend on exterior parameter – temperature. However, there is thermometer inside of the gyroscope. Therefore, during calibration and drawing up of the table we can take into account not only known angular velocity, but also the measured temperature [8-11]. By using such methods, the error of evaluation of the deviation considerably decreases. Nevertheless, the error has also some component, which cans changes for different experiments even at identical temperatures[1]. How can we reduce such error?

*2)* Separation the inconsistent properties over space.

It would be nice, if we can use some other similar gyroscope [4-6]. Then we would have two equations and two variables. However, it is forbidden by requirements of the problem. Indeed, it not completely so. The gyroscope has three axes, and only one of them is used for measuring rotation of a horizontal motion of the robot. The two other axes have no useful signal. However, the vibration and the inclination of the gyroscope axes (according to rotation axis or gravitation direction) create noise and projection of the useful signal on these axes also. Therefore, for each such axis we have one equation with two unknown values (the noise and the useful signal projection). We can include indications of these axes in calibration tables. By such a way, we can take into account the influence of vibrations and inclinations. I.e., "silent" axes are used as a measurement device of vibrations and inclinations of the axes of the gyroscope. We can cancel the error of angular velocity, depending on inclinations and frequencies, amplitudes of vibrations. By such a way, the error of evaluation of the noise considerably decreases.

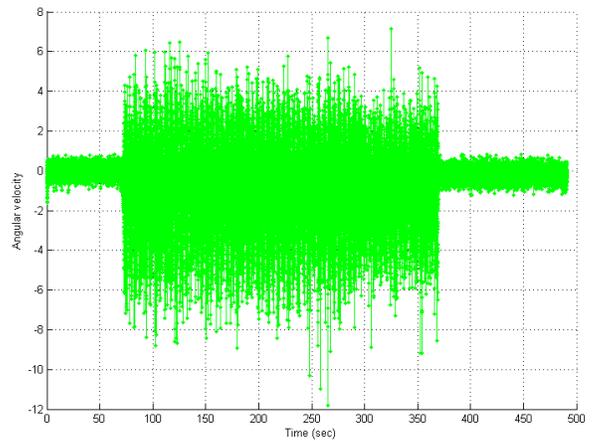

a)

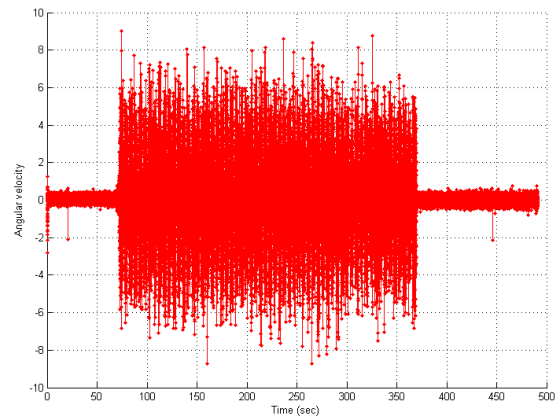

b)

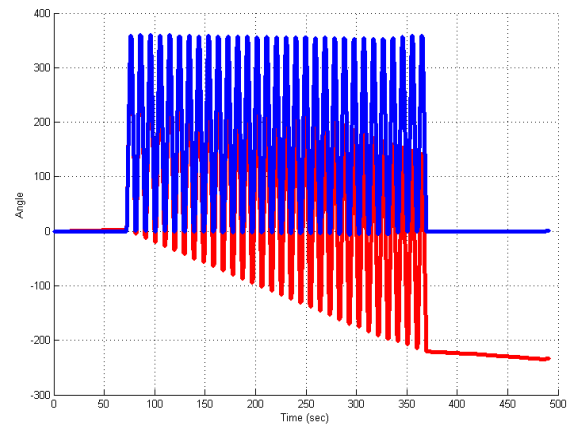

c)

Fig. 3. Output signals of the gyroscope on the robot (big vibrations) (a) $\omega_x^{(corr)}(t)$, (b) $\omega_y^{(corr)}(t)$, (c) $\varphi(t)$ – red line, $\varphi^{(corr)}(t)$ – blue line

---
[1] *In addition, the case is possible that error depends on factor that the device cannot measure - for example, electrical or magnetic field. In this case, we will use technical contradiction №2 (TC2):"We do not insert additional source of information. As a result, we get the simple device, but with the noise" We can transform TC2 to physical contradiction №2 (PC2): "The noise exists in the system. At the same time, the noise does not exist in the system." We can resolve PC2 by separation over space – to introduce screening surface against magnetic and electrical field.*

Also separation in Fourier space can be used. For example, using high frequencies amplitude, we can find value of vibrations. Using low frequencies components of motion, we can correct angular velocity by excluding some error, which is defined by found vibration.

Nevertheless, some random component remains, which can changes during one experiment. How can we reduce such random noise?

*E. Using standards*

  *1) Transition to subsystem 1*: Search of the missing information in subsystem

Subsystem is the indoor robot. Theoretically, his motion is arbitrary. Actually, for the real robot the motion is *smooth* [10-11, 13-18]. To be exact, the motion can described, mainly, into the following types:
a. No motion case
b. Straightforward motion with constant velocity
c. Rotation at constant angular velocity without straightforward motion
d. Rotation at constant angular acceleration without straightforward motion

From the form of the output signals, the minicomputer can "suggest" about type of smooth motion of the robot and this information gives us the additional information.

This mechanism can be implemented by the help of *Kalman filte* [18-24], which allows to separate random noise and a smooth motion. Thus, random noise and a smooth motion are described as linear mathematical models [18-24].

Using such methods, the random component of the error cam be considerably decreased, nevertheless, it remains. Integrating angular velocity to find angle, we summarize also this error, therefore the error of the angle increases in time. How can we prevent the error increase?

  *2) Transition to subsystem 2*: Search of the missing information in the global subsystem

The robot travels in a room. Rooms have walls, which are orthogonal to each other. The robot travels along these walls during most of the time. The analysis of the output signal can help to the minicomputer to suggest when the robot travels along one of these leading directions [25]. Thus, we can find absolute directions (defined with respect to no-motion system), instead of the relative directions. Reducing the error of the angle with respect to these absolute directions, we can correct the summarized error.

## IV. THE EXPERIMENTAL EXAMPLE

Let us consider the above-described procedure for reducing errors of some sample of gyroscope ST L3GD20 [12]. Let us consider a simple case when the temperature ($24^0$) and inclination of the gyro-sensor axes were the fixed. We consider Gyro sensor at two levels of vibrations – small and big. Small vibration exists on rotation table. Big vibration exists on indoor LG robot: Roboking.

Let $\omega_z^{(out)}$, $\omega_y^{(out)}$, $\omega_x^{(out)}$ be output angular velocities of the gyroscope.

Before the beginning of a motion, the gyroscope should be in no-motion state over 10 seconds. During this time, we detect average output signal over all three axes $\omega_{mean\ x}$, $\omega_{mean\ y}$, $\omega_{mean\ z}$ (in degree/sec).

We get from the experiment that $\omega_{mean\ z}$ is close to 1; $\omega_{mean\ y}$ is close to 0; $\omega_{mean\ x}$ is close to 1. However, these variables

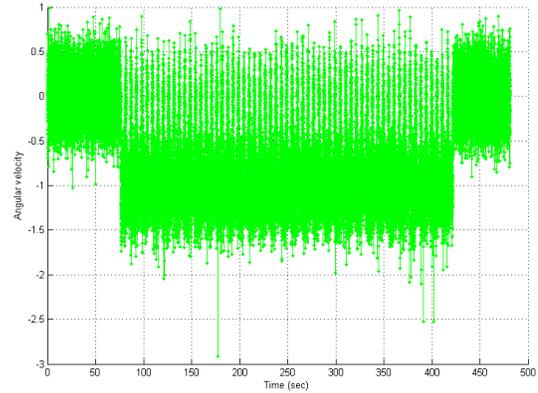

a)

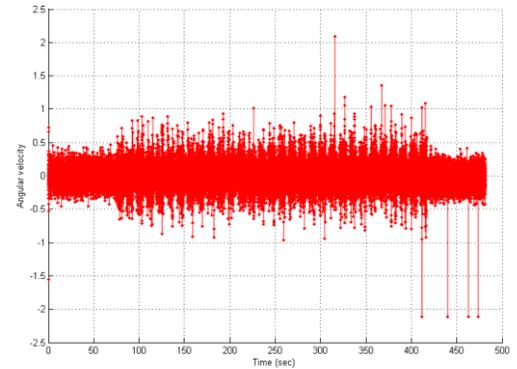

b)

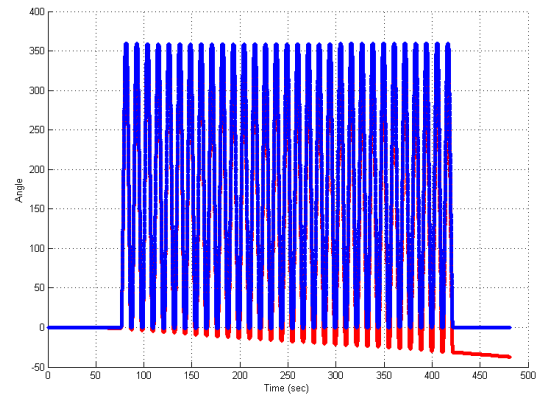

c)

Fig. 4. Output signals of the gyroscope on the rotation table (small vibrations) (a) $\omega_x^{(corr)}$ (t) , (b) $\omega_y^{(corr)}$ (t) , (c) φ (t) – red line, $φ^{(corr)}$(t) – blue line

can change its values for different experiments over 15 %. We correct the angular velocities over these variables:

$\omega_z = \omega_z^{(out)} - \omega_{mean\ z}$

$\omega_y = \omega_y^{(out)} - \omega_{mean\ y}$

$\omega_x = \omega_x^{(out)} - \omega_{mean\ x}$

For correction of random errors during no-motion, we use the method III.*E*.1). For correction of errors generated by vibration, we use the method III.*D*.2).

Let $\omega_z^{(corr)}$, $\omega_y^{(corr)}$, $\omega_x^{(corr)}$ be the angular velocities corrected by the above-mentioned methods

As the corrected rotation angle $\varphi^{(corr)}$ is following:

$\varphi^{(corr)}(0) = \varphi(0) = 0$

$\varphi(T) = \int_0^T dt \omega_z(t) = T_s \sum_i^N \omega_z(i)$

$\varphi^{(corr)}(T) = \int_0^T dt \omega_z^{(corr)}(t) = T_s \sum_i^N \omega_z^{(corr)}(i)$

where $T_s = 0.01$ s – an interval between measurements, $N = T/T_s$

Let's find now the updated angular velocities $\omega_z^{(corr)}$, $\omega_y^{(corr)}$, $\omega_x^{(corr)}$

For correction of random errors during no-motion, we use the method III.*E*.1).

For detection of no-motion conditions, we use the following requirements:

$|\omega_z| < 0.15/\sqrt{T_s}$

$|\omega_x| < 0.15/\sqrt{T_s}$

$|\omega_y| < 0.15/\sqrt{T_s}$

Indeed, velocity of rotation of a real robot cannot be too small. Therefore, we can observe such too small output signal only for no-motion case.

For no-motion conditions, we can assume that angular velocity of Gyro sensor is equal to zero:

$\omega_z^{(corr)} = 0$

Now let us consider the motion case. For correction of errors generated by vibration, we use the method III.*D*.2). Axis Z of the gyroscope cannot be aligned exactly vertically. Therefore, it is necessary to extract from $\omega_x$ and $\omega_y$ the projection of the angular velocity because of inclination:

$\omega_x^{(corr)} = \omega_x - \alpha_{xz} \omega_z$

$\omega_y^{(corr)} = \omega_y - \alpha_{yz} \omega_z$

where

$\alpha_{xz}$ is as coefficient of the linear regression between $\omega_x$ и $\omega_z$

$\alpha_{yz}$ is as coefficient of the linear regression between $\omega_y$ и $\omega_z$

The received numerical values the following:

*1) Rotation table:*

$\alpha_{xz} = -0.0074$

$\alpha_{yz} = -0.0013$

*2) Robot Roboking:*

$\alpha_{xz} = 0.0141$

$\alpha_{yz} = -0.0338$

Let us consider the following test for the robot motion:

The sensor motion consists of sequence of four steps:

- Left rotation on 360 degree (rotation velocity is equal to 90 degree/second)
- Delay of 0.35 second
- Right rotation on 360 degree (rotation velocity is equal to 90 degree/second)
- Delay of 0.35 second

These steps repeat 30 times

Let us give below graphs for output signals of the gyroscope: on the robot (Fig. 3) and on the rotation table (Fig. 4)

We can use the axes X and Y as measure device for vibration amplitude (in case of motion)

From Fig. 3 and Fig. 4 we can see that

*1)* The vibration results in random noise. The noise amplitude will increase if the vibration increases.

*2)* The systematic shift of angular velocity appears during vibration. However, this shift is almost independent on the vibration amplitude. The shift is approximately equal to $<\omega_x^{(corr)}> \approx -\omega_{mean\ x}$ and $<\omega_y^{(corr)}> \approx -\omega_{mean\ y}$ for axes X and Y correspondently.

It is visible, that the amplitude of random noise can be measure of vibration amplitude (vibr):

vibr $\approx <(\omega_x^{(corr)} - <\omega_x^{(corr)}>)^2>$

or

vibr $\approx <(\omega_y^{(corr)} - <\omega_y^{(corr)}>)^2>$

Experiment demonstrates that the correction that is necessary to compensate vibration noise, can be described by a constant bias. The constant bias is independent of current angular velocity and determined only by vibration, temperature, inclination and $\omega_{mean\ z}$.

Let us write it as the following formula:

a) *For no-motion conditions*:

$\omega_z^{(corr)} = 0$

b) *For the case of motion*:

$\omega_z^{(corr)} = \omega_z + k(temp, \omega_{mean\ z}, vibr, \alpha_{xz}, \alpha_{yz})$

*1) For big vibrations of the robot vibr_Roboking*:

$k(24^0C, \omega_{mean\ z}, vibr\_Roboking, \alpha_{xz}, \alpha_{yz}) = -0.37 + \omega_{mean\ z}$

a) *For no-motion conditions*:

$\omega_z^{(corr)} = 0$

b) *For the case of motion*:

$\omega_z^{(corr)} = \omega_z + k(temp, \omega_{mean\ z}, vibr, \alpha_{xz}, \alpha_{yz}) = \omega_z^{(out)} - 0.37$

*2) For small vibrations of the rotation table vibr_RT*:

$k(24^0C, \omega_{mean\ z}, vibr\_RT, \alpha_{xz}, \alpha_{yz}) = -1 + \omega_{mean\ z}$

a) *For no-motion conditions*:

$\omega_z^{(corr)} = 0$

b) *For the case of motion*:

$\omega_z^{(corr)} = \omega_z + k(temp, \omega_{mean\ z}, vibr, \alpha_{xz}, \alpha_{yz}) = \omega_z^{(out)} - 1$

Let us note that for small vibrations the final correction is -1 for output angular velocity $\omega_z^{(out)}$. It is close to $-\omega_{mean\ z}$.

## V. CONCLUSION

We solved the problem of the increase of the precision for MEMS Gyro sensors for indoor robots with horizontal motion. We can by using TRIZ's methods not only "develop" earlier-known algorithms, but also develop essentially new algorithms which eliminates noise, generated by vibrations, inclination and temperature. We developed the original algorithm when we use "silent" horizontal axes as measure devices of vibration and inclination. Indeed, we demonstrate that this algorithm can eliminate noise, generated by vibrations in real experiments (for fixed temperature and inclination).

ACKNOWLEDGMENT

We thank Kudryavtsev Alexander Vladimirovich for his very useful comments about this paper.